\documentclass[letterpaper]{article} % DO NOT CHANGE THIS
\usepackage[preprint]{aaai2027}  % DO NOT CHANGE THIS
\usepackage[hyphens]{url}  % DO NOT CHANGE THIS
\usepackage{graphicx} % DO NOT CHANGE THIS
\usepackage{natbib}  % DO NOT CHANGE THIS AND DO NOT ADD ANY OPTIONS TO IT
\usepackage{caption} % DO NOT CHANGE THIS AND DO NOT ADD ANY OPTIONS TO IT
\usepackage{algorithm}
\usepackage{algorithmic}
\usepackage{booktabs}
\usepackage{amsmath,amssymb,bm} % math + symbols + bold symbols
\usepackage{multirow}
\usepackage{array}
\usepackage{tabularx}
\usepackage{makecell}
\usepackage{subcaption}
\usepackage{xcolor}            % colored table rows
\usepackage{colortbl}
\usepackage{pifont}
\definecolor{mygray}{gray}{.88}

\newcommand{\cmark}{\ding{51}}
\newcommand{\xmark}{\ding{55}}
\definecolor{oursgreen}{RGB}{232,245,232}
\definecolor{lightgray}{gray}{0.92}
\title{GestureFAR: Streaming Co-Speech Gesture Generation with Flow Autoregression}

\author{
    Pinxin Liu$^{1}$, Haiyang Liu$^{2}$, Jiahao Luo$^{3}$, Junhua Huang$^{4}$, Chunhao Zou$^{5}$, Luchuan Song $^{1}$
}
\affiliations{
    $^{1}$University of Rochester, $^{2}$University of Tokyo, $^{3}$University of California Santa Cruz, 
    
    $^{4}$University of California Los Angeles, $^{5}$Meta
}

\begin{document}

\twocolumn[{
\renewcommand\twocolumn[1][]{#1}%
\maketitle
\begin{center}
    \centering
    \captionsetup{type=figure}
    \vspace{-1cm}
    \includegraphics[width=\linewidth]{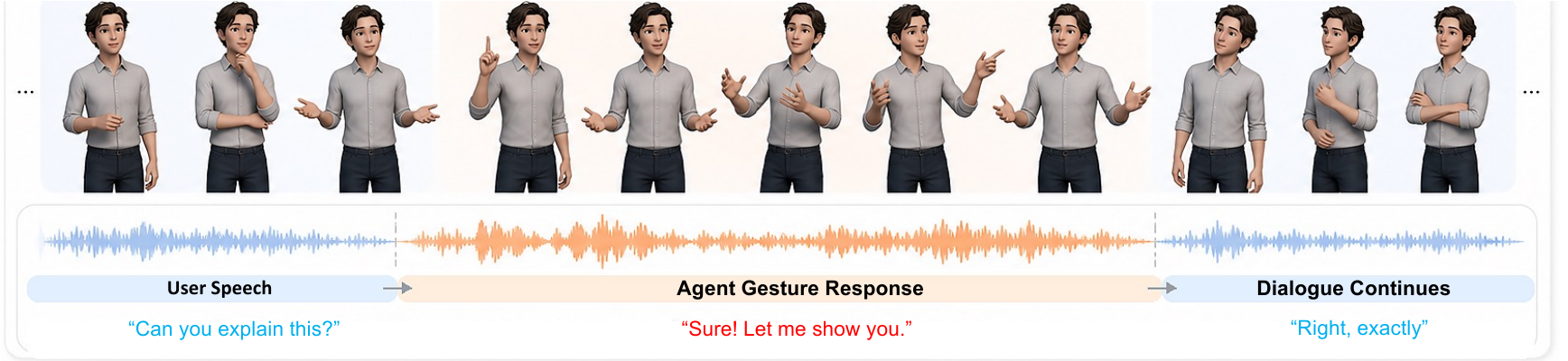}
    \caption{
        GestureFAR enables live interactive co-speech gesture generation: as user speech arrives incrementally, the embodied agent keeps producing synchronized gesture responses throughout the ongoing dialogue.
    } \label{fig:teaser}
    % \vspace{-0.3cm}
\end{center}
}]

% \maketitle

% \begin{figure*}[t]
% \centering
% \includegraphics[width=\textwidth]{figs/gesturemar_teaser.pdf}
% \caption{GestureFAR enables live interactive co-speech gesture generation: as user speech arrives incrementally, the embodied agent keeps producing synchronized gesture responses throughout the ongoing dialogue.}
% \label{fig:teaser}
% \end{figure*}

\begin{abstract}
Generating natural co-speech gestures from streaming speech is essential for embodied conversational agents, where motion must be produced while a user is still speaking. Recent streaming gesture systems make online generation possible by autoregressing over discrete motion tokens, but this design compresses high-dimensional continuous motion into finite codebooks and can limit the realism and diversity of generated gestures. To preserve both causality and continuous expressiveness, we propose \textbf{GestureFAR}, a flow-autoregressive framework for streaming co-speech gesture generation. First, GestureFAR autoregresses over causal continuous motion latents, using a transformer to model streaming audio-motion context and a per-token flow-matching head to sample the next latent from a continuous distribution. Second, we introduce a head-only flow distillation strategy that freezes the causal backbone and distills the multi-step per-token flow head into a single network evaluation using consistency and distribution-matching objectives. This keeps the model token-causal while removing the main latency bottleneck for live interaction. Experiments on BEAT2 show that GestureFAR significantly improves the quality--latency trade-off among streaming-capable methods, preserving strong gesture quality while enabling real-time token-causal generation. Project Page: {\small\url{https://andypinxinliu.github.io/GestureFAR}}
\end{abstract}


\section{Introduction}
\label{sec:intro}

Speech and gesture are tightly coupled in human communication. Alongside words, people use hand, arm, and body movements to express emotion, emphasize intent, structure discourse, and make conversations more engaging~\cite{de2012interplay,song2023emotional}. Generating such co-speech gestures is therefore an important capability for virtual avatars and embodied conversational agents~\cite{liu2023emage,yi2022generating}. As these agents move from offline animation to live interaction, gesture generation must also become streaming: the system should produce motion as speech arrives, without seeing future audio or revising previously emitted frames.

Recent co-speech gesture systems have achieved strong progress in realism and efficiency~\cite{liu2023emage,chen2024syntalker,liu2025gesturelsm}, but most are still designed for offline or block-based generation. They process a complete utterance, or at least a fixed audio window, before generating the corresponding motion. This setting is fundamentally different from live interaction, where an avatar must commit to each gesture using only past motion and currently available speech. Recent online systems such as MIBURI~\cite{mughal2026miburi} and LiveGesture~\cite{saleem2026livegesture} address this requirement with autoregressive decoding over discrete motion tokens, making strict streaming possible.

However, discrete tokenization introduces a new bottleneck. Full-body gestures are continuous, high-dimensional, and often distinguished by subtle changes in wrist rotation, arm trajectory, torso timing, and whole-body coordination. Compressing this space into a finite codebook can discard fine-grained motion details and impose a representation ceiling before the sequence model is trained. Continuous generative models such as diffusion and flow matching are better suited to such multimodal motion distributions, but standard diffusion-style generators are iterative and typically operate over full sequences or fixed blocks. 
To resolve this issue, we propose \textit{\textbf{GestureFAR}}, a streaming co-speech gesture framework based on \emph{flow autoregression}. GestureFAR keeps the left-to-right factorization of autoregressive models, but replaces discrete token prediction with continuous latent sampling. A causal motion VAE maps full-body motion into streamable continuous latents. A causal audio-motion transformer summarizes past motion and currently observed speech into a per-token condition. Instead of predicting a codebook index, a lightweight flow-matching head samples the next continuous latent from this condition. This design combines the token-causal emission of autoregressive systems with the continuous expressiveness of flow-based generative modeling.

% Thus, streaming gesture generation faces a core trade-off: discrete autoregressive models are token-causal but lose continuous expressiveness, while continuous generative models are expressive but difficult to deploy as strict left-to-right streamers.

Although the autoregressive backbone is causal, a multi-step flow sampler would still require repeated head evaluations for every emitted latent. We observe that this cost is localized in the flow head: the transformer has already produced the streaming audio-motion condition, while the head only performs local continuous sampling. Based on this separation, we introduce \emph{Multi-Procedure Distribution Matching Distillation}, a head-only distillation strategy that freezes the tokenizer and autoregressive backbone, caches their conditioning vectors, and distills only the per-token flow head into a one-step sampler. This makes continuous autoregressive gesture generation practical for live streaming.

\begin{figure}[]
\centering
  \includegraphics[width=.9\columnwidth, trim={0cm 0cm 0cm 0cm}, clip]{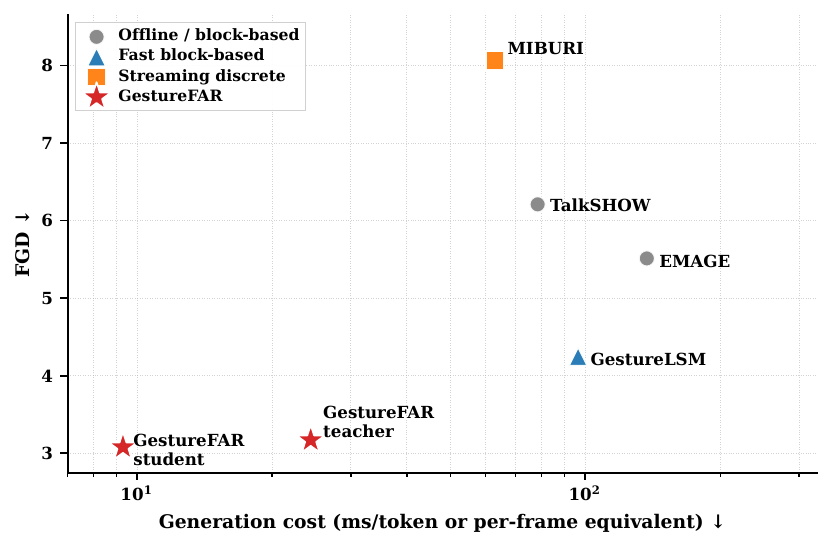}
\vspace{-0.3cm}
\caption{GestureFAR student lies in the lower-left region, achieving the best FGD with the lowest generation cost while remaining streamable and token-causal. 
}
\vspace{-5mm}
\label{fig:speed}
\end{figure}

Experiments on BEAT2 show that GestureFAR achieves a strong quality--latency trade-off among streaming-capable gesture systems. As visualized in Figure.~\ref{fig:speed}, GestureFAR better preserves continuous full-body motion quality while enabling real-time token-causal deployment. In summary, our contributions are:
\begin{itemize}
    \item We introduce GestureFAR, a streaming token-causal co-speech gesture framework that autoregresses over continuous motion latents instead of discrete gesture tokens.
    \item We design a flow-autoregressive generator in which a causal audio-motion transformer models streaming context and a per-token flow head samples the next latent.
    \item We propose Multi-Procedure Distribution Matching Distillation, a head-only one-step distillation strategy that freezes the causal backbone and distills only the flow head for live deployment.
\end{itemize}

\section{Related Work}\label{sec:related}

\paragraph{Co-Speech Gesture Generation.}
Existing works on co-speech gesture generation mostly employ skeleton- or joint-level pose representations~\cite{liu2022learning,liu2024tango,liu2025contextualgesturecospeechgesture,liu2025semgessemanticsawarecospeechgesture,song2023emotional}. HA2G~\cite{liu2022learning} constructs high- and low-level audio-motion embeddings for hierarchical gesture generation. TalkShow~\cite{yi2022generating} estimates SMPL-X~\cite{SMPL-X:2019} poses and jointly models body and hand motions for talk-show videos. CaMN~\cite{liu2022beat} and EMAGE~\cite{liu2023emage} introduce large-scale conversational datasets and a GPT-style decoder for joint face and body modeling with diverse style control. ProbTalk~\cite{probtalk} models gesture in a VQ token space. More recent works push for either expressive quality or inference speed: MambaTalk~\cite{mambatalk} accelerates generation through an efficient Mamba architecture, while DiffSHEG~\cite{diffsheg} and SynTalker~\cite{chen2024syntalker} adopt diffusion-based pipelines. GestureLSM~\cite{liu2025gesturelsm} introduces latent shortcut flow matching for few-step generation. None of these works targets strict token-level streaming inference, the regime we address.

\vspace{-0.1cm}
\paragraph{Streaming and Causal Motion Generation.}
Streaming motion generation has received limited attention. RDLA~\cite{vu2026streaming} and FloodDiffusion~\cite{cai2026flooddiffusion} propose streaming diffusion variants for gesture, while Diffusion Forcing~\cite{chen2024diffusionforcing} provides a general AR-style rolling diffusion framework. These methods preserve temporal causality but still rely on iterative denoising. MIBURI~\cite{mughal2026miburi} and LiveGesture~\cite{saleem2026livegesture} instead achieve online gesture generation with discrete autoregressive tokens. Their token-causal formulation is well suited to low-latency decoding, but the discrete tokenizer compresses continuous body motion into finite codebooks, introducing a representation ceiling for fine-grained gesture generation. GestureFAR is inspired by continuous-latent autoregressive generation in images and videos~\cite{li2024autoregressiveimage,deng2025autoregressivevideo}, but adapts the idea to streaming co-speech gesture: we use a strictly causal AR transformer for audio-motion context and generate each continuous latent token with a rectified-flow head.

\vspace{-0.1cm}
\paragraph{Few-Step Flow Models and Diffusion Distillation.}
Diffusion and flow-based models offer strong generative quality, but their iterative solvers are expensive for live streaming. Flow matching~\cite{lipman2022flow} and rectified flow~\cite{liu2022flowstraightfastlearning} learn an ODE between noise and data, while shortcut models~\cite{shortcutmodels} train a single network to predict flow shortcuts at multiple step sizes. A parallel line accelerates diffusion sampling through distillation: consistency models~\cite{song2023consistency} map noisy states directly to clean endpoints, score-regularized continuous-time variants~\cite{zheng2026rcm} stabilize the objective with forward-mode derivatives, and distribution-matching distillation~\cite{yin2024one} improves sample quality through an auxiliary score network. GestureFAR uses a multi-step rectified-flow head as the teacher and distills it to one evaluation per emitted motion token, making continuous-latent autoregression practical for real-time streaming.

\section{Method}
\label{sec:method}

\begin{figure*}[t]
\centering
\includegraphics[width=2\columnwidth]{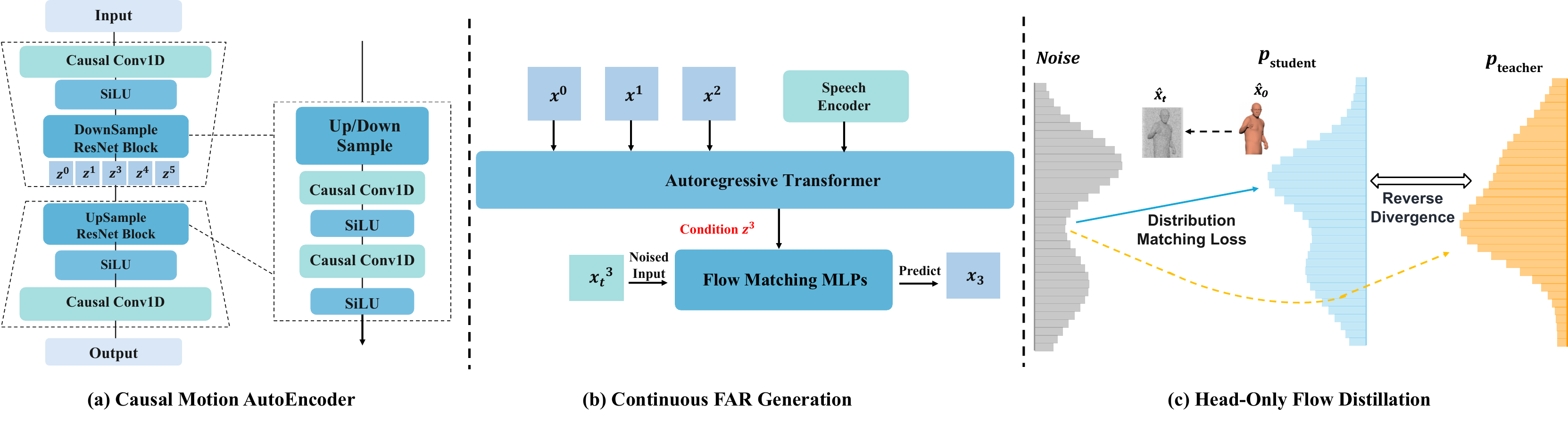}
\caption{
\textbf{Overview of GestureFAR.}
GestureFAR generates co-speech gestures from streaming audio by autoregressing over continuous motion latents. A causal motion VAE maps full-body motion into a streamable latent representation. A streaming audio-motion transformer predicts the condition for each next latent using past motion and currently available audio. A lightweight per-token flow head samples the next continuous latent from this condition. For real-time deployment, we distill only the flow head into a one-step student while keeping the transformer frozen.
}
\label{fig:pipeline}
\end{figure*}

As shown in Fig.~\ref{fig:pipeline}, GestureFAR is designed to emit gestures causally without reducing continuous body motion to discrete codebook prediction. We first present the flow-autoregressive framework in Sec.~\ref{sec:method:gesturefar}, including the causal motion VAE, streaming audio-motion transformer, and continuous flow head. We then introduce head-only one-step distillation in Sec.~\ref{sec:method:distill}, which removes the multi-step sampling cost while keeping the causal transformer fixed. Finally, Sec.~\ref{sec:method:streaming} describes the stateful streaming deployment loop.

\subsection{GestureFAR Framework}
\label{sec:method:gesturefar}

Streaming gesture generation imposes three coupled requirements. First, the motion representation itself must be causal; otherwise the generator can inherit future information from the tokenizer. Second, each emitted token must depend only on past motion and available speech. Third, the token distribution should remain continuous and multimodal, since a single speech segment can correspond to multiple plausible gestures. Let $\mathbf{z}_{1:N}$ be the motion latents produced by the VAE and $\mathbf{h}_{\le i}$ be the speech features available before latent $i$ is emitted. GestureFAR encodes the streaming constraint as
\begin{equation}
p(\mathbf{z}_{1:N}\mid\mathbf{a}_{1:T})
=
\prod_{i=1}^{N}
p_{\theta,\phi}
\left(
\mathbf{z}_i
\mid
\mathbf{z}_{<i},
\mathbf{h}_{\le i}
\right),
\label{eq:streaming-factorization}
\end{equation}
and implements each factor with the three modules below.

\paragraph{Causal Continuous Motion Tokenizer.}
The first module addresses representation. A non-causal tokenizer is incompatible with streaming because the latent at step $i$ may already contain future frames. A discrete tokenizer is streamable, but it can cap gesture quality by forcing subtle hand and upper-body variations into a finite codebook. We therefore use a 1D causal motion VAE that keeps the latent space continuous while preventing future leakage. Given full-body motion $\mathbf{m}_{1:T}$, the encoder $E$ produces a temporally compressed latent sequence
% \begin{equation}
% \boldsymbol{\mu}_{1:N},\boldsymbol{\sigma}_{1:N}=E(\mathbf{m}_{1:T}),
% \qquad
% \mathbf{z}_i=\boldsymbol{\mu}_i+\boldsymbol{\sigma}_i\odot\boldsymbol{\epsilon}_i,
% \label{eq:vae-encode}
% \end{equation}
and the decoder $D$ reconstructs it.
% $\hat{\mathbf{m}}_{1:T}=D(\mathbf{z}_{1:N})$. 
All temporal convolutions in $E$ and $D$ are causal, so $\mathbf{z}_i$ and the reconstructed frames emitted from it depend only on motion frames up to the corresponding time. 
%
% We train the causal VAE with a standard reconstruction-regularization objective,
% \begin{equation}
% \begin{aligned}
% \mathcal{L}_{\mathrm{VAE}}
% \;=&\;
% \|\mathbf{m}_{1:T}-\hat{\mathbf{m}}_{1:T}\|_1 \\
% &+
% \lambda_{\mathrm{KL}}
% D_{\mathrm{KL}}
% \left(
% q_E(\mathbf{z}\mid\mathbf{m})
% \;\|\;
% \mathcal{N}(\mathbf{0},\mathbf{I})
% \right).
% \end{aligned}
% \label{eq:vae-loss}
% \end{equation}
% This tokenizer plays the same role as discrete motion tokens in autoregressive streaming systems, but without quantizing away fine-grained motion. 
At streaming inference, its convolution states are cached across chunks, making both encoding and decoding incremental.

\paragraph{Streaming Audio-Motion Transformer.}
The second module addresses conditioning. Once the motion is represented as causal latents, the model must decide what information is available when predicting the next latent. GestureFAR uses a streaming transformer to summarize previous motion and currently available speech into a condition vector:
\begin{equation}
\mathbf{c}_i=f_\theta(\mathbf{z}_{<i},\mathbf{h}_{\le i}),
\qquad
\mathbf{z}_i\sim p_\phi(\mathbf{z}\mid\mathbf{c}_i).
\label{eq:condition-vector}
\end{equation}
The first term produces the causal context, and the second term is generated by the flow head. Each transformer block applies causal self-attention over motion latents followed by causal cross-attention to speech features. For latent $i$, motion self-attention can only see $\mathbf{z}_{<i}$, and audio cross-attention can only see $\mathbf{h}_{\le i}$. Thus the transformer is responsible for continuity and speech alignment, but it does not collapse the next motion into a deterministic prediction.

\paragraph{Continuous Flow Head.}
The third module addresses generation. A deterministic regressor would average over plausible gestures, while a discrete classifier would return to the codebook bottleneck we want to avoid. Inspired by continuous autoregressive generation, we attach a lightweight MLP flow head to each transformer condition. The head operates on one latent token at a time: the transformer provides $\mathbf{c}_i$, and the head models the conditional continuous distribution of $\mathbf{z}_i$. For a clean latent $\mathbf{z}_i$, noise $\boldsymbol{\epsilon}\sim\mathcal{N}(\mathbf{0},\mathbf{I})$, and flow time $\tau\in[0,1]$, define
\begin{equation}
\mathbf{x}_{\tau}
=
(1-\tau)\mathbf{z}_i    
+
\tau\boldsymbol{\epsilon},
\qquad
\mathbf{u}_{\tau}
=
\boldsymbol{\epsilon}-\mathbf{z}_i .
\label{eq:flow-path}
\end{equation}
The teacher head $\mathbf{v}_{\phi}$ is trained by flow matching:
\begin{equation}
\mathcal{L}_{\mathrm{FM}}
=
\mathbb{E}_{i,\tau,\boldsymbol{\epsilon}}
\left[
\left\|
\mathbf{v}_{\phi}
(
\mathbf{x}_{\tau},\tau;\mathbf{c}_i
)
-
\mathbf{u}_{\tau}
\right\|_2^2
\right].
\label{eq:flow-loss}
\end{equation}
At inference, the teacher samples $\mathbf{z}_i$ by integrating this velocity field from noise to data. This gives a strong continuous sampler, but it requires multiple head evaluations for every emitted latent. The next section removes this cost.

\subsection{Multi-Procedure Distribution Matching Distillation}
\label{sec:method:distill}

The main acceleration opportunity in GestureFAR is that the expensive multi-step computation is confined to the flow head. The streaming transformer has already produced the causal audio-motion condition; distilling the entire transformer would be expensive and unnecessary. We therefore freeze the tokenizer and AR transformer, cache their conditioning vectors $\mathbf{c}_t$, and distill only the flow head:
\begin{equation}
p_{\phi}(\mathbf{z}_t\mid \mathbf{c}_t)
\;\rightarrow\;
p_{\psi}(\mathbf{z}_t\mid \mathbf{c}_t),
\label{eq:head-distill}
\end{equation}
where $\phi$ is the multi-step teacher head and $\psi$ is the one-step student head. This turns distillation from an end-to-end sequence-model problem into a compact conditional sampling problem. It is faster to train, more stable, and preserves the original causal factorization.

Let $G_{\psi}(\boldsymbol{\epsilon},\mathbf{c})$ denote the one-step student sample, and let $F_{\psi}(\mathbf{x}_{\tau},\tau;\mathbf{c})$ denote the student's endpoint prediction from an intermediate state. We train the student with two complementary procedures.

\paragraph{Discrete consistency warm-up.}
We first initialize the student with teacher solver endpoints. Given an intermediate noisy latent $\mathbf{x}_{\tau}$, we integrate the teacher flow head toward the data endpoint,
\begin{equation}
\mathbf{x}_{0}^{\phi}
=
\mathrm{ODE}
\left(
\mathbf{x}_{\tau},\tau\rightarrow 0;
\mathbf{v}_{\phi},\mathbf{c}
\right),
\label{eq:teacher-endpoint}
\end{equation}
and train the student endpoint prediction to match this stop-gradient target:
\begin{equation}
\mathcal{L}_{\mathrm{dCM}}
=
\mathbb{E}
\left[
\left\|
F_{\psi}(\mathbf{x}_{\tau},\tau;\mathbf{c})
-
\mathrm{sg}[\mathbf{x}_{0}^{\phi}]
\right\|_2^2
\right].
\label{eq:dcm}
\end{equation}
This stage gives a stable one-step initialization, although it inherits the discretization error of the teacher solver.

% \paragraph{Continuous consistency refinement.}
% We then refine the student by enforcing that its endpoint prediction is constant along the teacher probability-flow trajectory. Intuitively, if two noisy states lie on the same teacher trajectory, the student should map both to the same clean latent. Let the teacher trajectory follow $d\mathbf{x}_{\tau}/d\tau=\mathbf{v}_{\phi}(\mathbf{x}_{\tau},\tau;\mathbf{c})$. The ideal endpoint map satisfies
% \begin{equation}
% \frac{d}{d\tau}
% F_{\psi}(\mathbf{x}_{\tau},\tau;\mathbf{c})
% =
% \partial_{\tau}F_{\psi}
% +
% J_{\mathbf{x}}F_{\psi}\,
% \mathbf{v}_{\phi}
% =0.
% \label{eq:scm}
% \end{equation}
% We minimize the squared norm of this residual. This continuous objective removes the solver-discretization ceiling of the warm-up stage and better aligns the one-step map with the teacher dynamics.

\paragraph{Distribution matching.}
Consistency aligns the student to the teacher trajectory, but it can still under-cover the conditional distribution. To preserve multimodality, we add a DMD-style~\cite{yin2024one} distribution matching term. We draw a one-step student sample and re-noise it,
\begin{equation}
\hat{\mathbf{z}}
=
G_{\psi}(\boldsymbol{\epsilon},\mathbf{c}),
\qquad
\mathbf{y}_{\sigma}
=
\alpha_{\sigma}\hat{\mathbf{z}}
+
\sigma\boldsymbol{\eta},
\label{eq:dmd-perturb}
\end{equation}
where $\boldsymbol{\eta}\sim\mathcal{N}(\mathbf{0},\mathbf{I})$. We compare the teacher score $s_{\phi}$ with a fake score $s_{\omega}$ trained on student samples and update the student with the score-gap gradient
\begin{equation}
\begin{aligned}
\nabla_{\psi}\mathcal{L}_{\mathrm{DMD}}
\;=\;&
\mathbb{E}
\left[
\lambda(\sigma)
\left(
s_{\omega}(\mathbf{y}_{\sigma},\sigma;\mathbf{c})
-
s_{\phi}(\mathbf{y}_{\sigma},\sigma;\mathbf{c})
\right) \right. \\
&\left.
\frac{\partial G_{\psi}(\boldsymbol{\epsilon},\mathbf{c})}{\partial \psi}
\right].
\end{aligned}
\label{eq:dmd}
\end{equation}
The fake score is updated by denoising score matching. This term helps the one-step head preserve the diversity of the multi-step teacher. The final distillation objective is
\begin{equation}
\mathcal{L}_{\mathrm{distill}}
=
\mathcal{L}_{\mathrm{dCM}}
% +
% \lambda_{\mathrm{sCM}}
% \mathcal{L}_{\mathrm{sCM}}
+
\lambda_{\mathrm{DMD}}
\mathcal{L}_{\mathrm{DMD}}.
\label{eq:distill-total}
\end{equation}
% After distillation, the student head replaces the teacher head. The backbone and tokenizer are unchanged, so the model remains strictly causal while requiring only one head evaluation per latent.

\begin{figure}[]
% \vspace{-3mm}
\centering
  \includegraphics[width=1\columnwidth, trim={0cm 0cm 0cm 0cm}, clip]{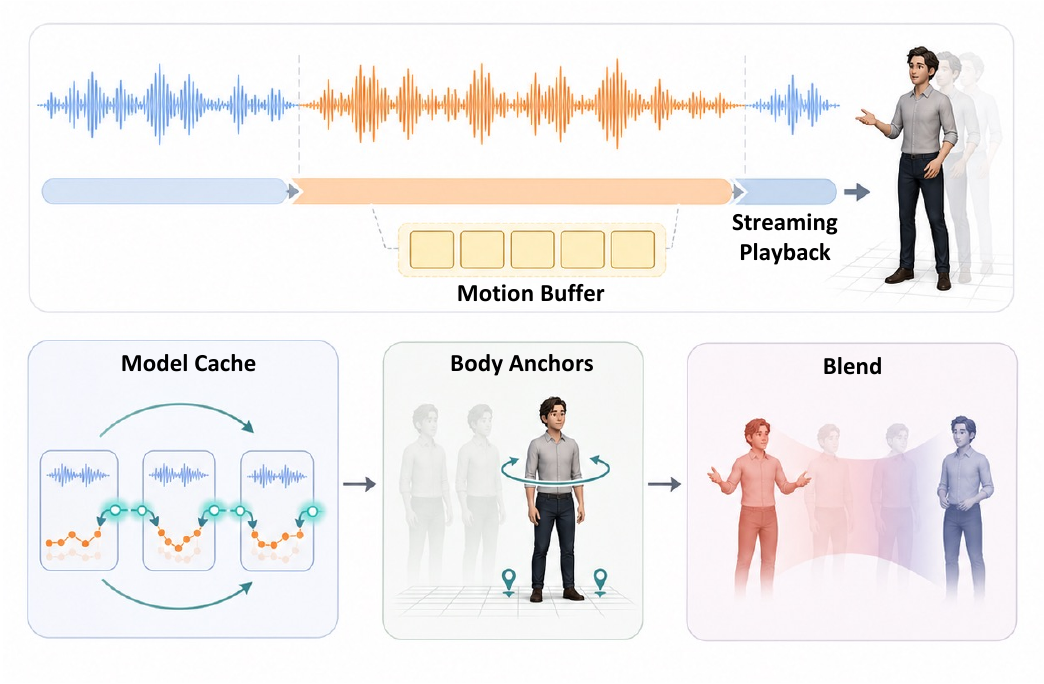}
\vspace{-3mm}

\vspace{-0.5cm}
\caption{\textbf{Stateful streaming deployment.} GestureFAR is deployed with persistent model caches, body anchors, a short motion buffer, and motion-space blending. The caches make chunked generation incremental, body anchors prevent translation drift and re-centering artifacts, and the buffer decouples bursty backend generation from fixed-rate frontend playback. Blending smooths transitions between speaking, listening, and idle states, enabling stable live avatar rendering.}

\label{fig:deploy}
\end{figure}

\subsection{Stateful Streaming Deployment}
\label{sec:method:streaming}

As shown in Fig.~\ref{fig:deploy}, at deployment, GestureFAR is not executed as a sequence of independent audio chunks. A live avatar must handle irregular audio arrival, bursty GPU execution, fixed-rate rendering, and pauses between speaking turns. We therefore use a stateful streaming runtime. For chunk $k$, the runtime emits newly available motion frames while updating a persistent state:
\begin{equation}
\mathbf{o}_k,\,\mathbf{s}_{k+1}
=
\mathcal{G}_{\psi}(\mathbf{a}_k,\mathbf{s}_k).
\label{eq:streaming-state}
\end{equation}
Here $\mathbf{s}_k$ contains the audio encoder cache, autoregressive transformer cache, causal VAE decoder cache, and body anchors for translation, floor height, and facing direction. These anchors attach each decoded chunk to the previously rendered body state, preventing chunk-wise re-centering from appearing as body snapping, vertical drifting, or sudden reorientation.

The runtime also decouples generation from display with a short motion buffer. The backend produces frames in chunk-level bursts, while the renderer consumes them at a fixed frame rate. This producer-consumer design absorbs variations caused by chunk boundaries, GPU scheduling, and rendering load. Keeping the buffer short bounds response latency, while maintaining enough buffered frames avoids visible stalls during live playback.

Finally, we treat speaking and listening as runtime states rather than separate learned modes. During speaking, audio-conditioned GestureFAR motion is generated and the last rendered pose is stored. During listening, when no speech-conditioned motion is available, the runtime loops a short idle motion clip through the same retargeting and playback buffer, so the avatar remains alive without hallucinating speech gestures. We smooth each state transition with a short motion-space blend:
\begin{equation}
\tilde{\mathbf{m}}_j
=
\mathrm{Blend}(\mathbf{m}^{A}_j,\mathbf{m}^{B}_j;\alpha_j),
\label{eq:runtime-blend}
\end{equation}
where rotations are interpolated by quaternion slerp and translations by linear interpolation. The same blending is used for speaking-to-listening, listening-to-speaking, and idle-loop wraparound, with idle translation pinned to the last speaking position to avoid teleportation.

\begin{figure*}[t]
    \centering%pipeline_crop
    \includegraphics[width=1\linewidth]{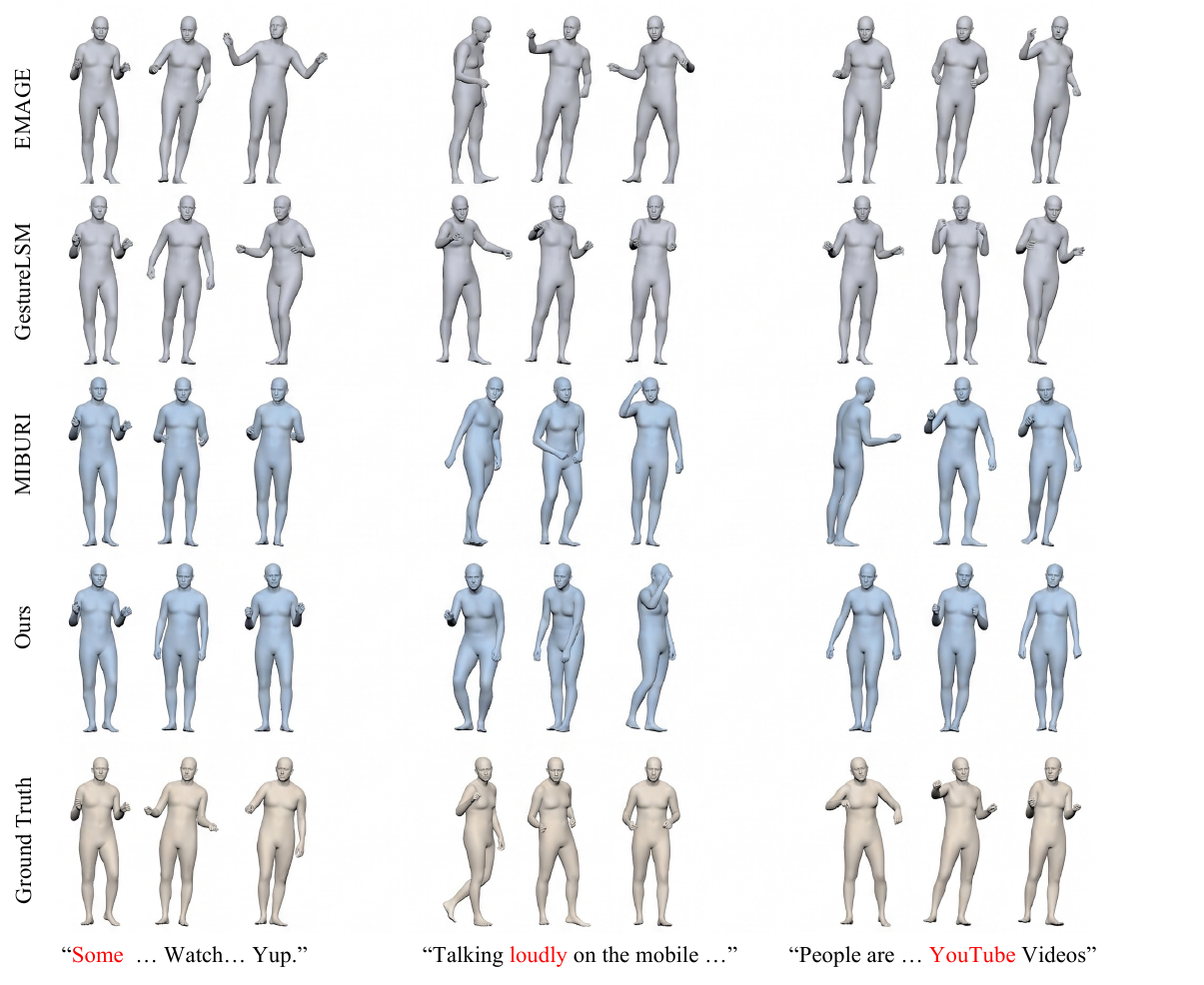}
    \vspace{-0.3cm}
    \caption{\textbf{Qualitative comparison on BEAT2.} We visualize generated full-body gestures from different methods under the same speech inputs, with the semantically salient words highlighted in red. Compared with prior methods, our model produces gestures that are more natural, diverse, and better aligned with the speech semantics.} \label{fig:visualization}
    \vspace{-0.5cm}
\end{figure*}

\begin{table*}[t]
\caption{
\textbf{State-of-the-art comparison on BEAT2.}
Offline and block-based baselines follow the LiveGesture comparison protocol. 
We additionally report whether each method supports streaming generation and strict token-causal emission. 
FGD is lower better. BC and Diversity are marked with $\rightarrow$ because values closer to the ground-truth statistics are preferred. 
Runtime is reported as per-token generation cost when available.
}
\label{tab:beat2-main}
\centering
\small
\setlength{\tabcolsep}{3.8pt}
\resizebox{\textwidth}{!}{%
\begin{tabular}{l l c c c c c c}
\toprule
\textbf{Method} 
& \textbf{Venue} 
& \textbf{Streaming} 
& \makecell{\textbf{Token}\\\textbf{causal}} 
& \textbf{FGD $\downarrow$} 
& \textbf{BC $\rightarrow$} 
& \textbf{Diversity $\rightarrow$} 
& \textbf{ms/token $\downarrow$} \\
\midrule
\rowcolor{gray!10}
\multicolumn{8}{c}{\textbf{Offline / block-based solutions}} \\
\midrule
CaMN~\cite{liu2022beat}                    
& ECCV'22    
& \xmark & \xmark 
& 6.644 & 0.676 & \underline{10.86} & -- \\

TalkSHOW~\cite{yi2022generating}           
& CVPR'23    
& \xmark & \xmark 
& 6.209 & \textbf{0.695} & 13.47 & -- \\

ProbTalk~\cite{probtalk}                   
& CVPR'24    
& \xmark & \xmark 
& 5.040 & 0.771 & 13.27 & -- \\

EMAGE~\cite{liu2023emage}                  
& CVPR'24    
& \xmark & \xmark 
& 5.512 & 0.772 & 13.06 & -- \\

MambaTalk~\cite{mambatalk}                 
& NeurIPS'24 
& \xmark & \xmark 
& 5.366 & 0.781 & 13.05 & -- \\

SynTalker~\cite{chen2024syntalker}         
& ACMMM'24      
& \xmark & \xmark 
& 4.687 & 0.736 & \textbf{12.43} & -- \\

GestureLSM~\cite{liu2025gesturelsm}        
& ICCV'25    
& \xmark & \xmark 
& 4.247 & \underline{0.729} & 13.76 & -- \\
\midrule
\rowcolor{gray!10}
\multicolumn{8}{c}{\textbf{Streaming solutions}} \\
\midrule
MIBURI~\cite{mughal2026miburi}             
& CVPR'26    
& \cmark & \cmark 
& 8.06 & 0.790 & 17.50 & 62.9 \\

LiveGesture~\cite{saleem2026livegesture}   
& CVPR'26    
& \cmark & \cmark 
& 4.57 & 0.794 & 13.91 & -- \\

GestureFAR teacher, $8$-step               
& Ours       
& \cmark & \cmark 
& \underline{3.17} & 0.734 & 13.31 & 24.4 \\

\rowcolor{oursgreen}
\textbf{GestureFAR student, $1$-step}       
& Ours       
& \cmark & \cmark 
& \textbf{3.08} & 0.741 & 13.24 & \textbf{9.3} \\
\midrule
\rowcolor{gray!10}
Ground Truth                                
& --         
& -- & -- 
& -- & 0.703 & 11.97 & -- \\
\bottomrule
\end{tabular}
}
\end{table*}

\section{Experiments}
\label{sec:exp}

\subsection{Experimental Setup}
\label{sec:exp:setup}

\paragraph{Dataset.}
We train and evaluate on BEAT2~\cite{liu2022beat,liu2023emage}, a standard full-body co-speech gesture benchmark with approximately $60$ hours of SMPL-X motion and speech from $25$ speakers over $1{,}762$ conversational clips. Following recent BEAT2 evaluations~\cite{liu2023emage,yi2022generating,mambatalk,chen2024syntalker,liu2025gesturelsm}, we report the main results on the speaker-$2$ test split and use the official train/validation/test protocol.

\paragraph{Metrics.}
We evaluate gesture realism using FGD~\cite{yoon2020speech}, speech--motion synchronization using Beat Consistency (BC)~\cite{li2021ai}, and motion variation using L1 Diversity (Div.)~\cite{li2021audio2gestures}. Since GestureFAR targets online deployment, we additionally mark whether a method is \emph{streaming} and whether it is \emph{token-causal}, and report per-token generation cost (ms/token) and real-time factor (RTF) when available. A streaming method runs without seeing the complete utterance, while a token-causal method emits each motion token without future audio or future motion.

\paragraph{Implementation details.}
GestureFAR uses a causal motion VAE with latent dimension $128$ and temporal downsampling stride $4$, producing motion latents at $7.5$Hz. The autoregressive backbone is a $8$-layer causal transformer with hidden size $384$, $6$ attention heads, and FFN dimension $1024$. Audio features are extracted with a frozen mHuBERT encoder. The per-token flow head is a $4$-layer SimpleMLPAdaLN. The teacher is sampled with an $8$-step Heun solver; the deployment model is a one-step distilled student. During distillation, the autoregressive backbone is frozen and only the per-token flow head is trained on cached conditioning--target pairs.

\subsection{Evaluation Results}

\label{sec:exp:main}

Figure~\ref{fig:visualization} shows qualitative comparisons under the same speech inputs. GestureFAR produces more natural and speech-relevant gestures, with expressive arm and hand movements on emphasized words such as ``Some,'' ``loudly,'' and ``YouTube,'' while preserving plausible body posture and temporal continuity. In contrast, prior methods often generate over-smoothed motion, weaker speech-gesture correspondence, or unnatural body configurations. Table~\ref{tab:beat2-main} reports quantitative results on BEAT2, following the comparison protocol of recent full-body gesture systems and including the closest streaming methods, MIBURI~\cite{mughal2026miburi} and LiveGesture~\cite{saleem2026livegesture}. Most prior high-quality methods are offline or block-based, while existing streaming methods rely on discrete autoregressive tokens. GestureFAR instead performs token-causal generation over continuous motion latents, achieving the best FGD among all compared methods. The one-step student remains ahead of both offline and streaming baselines while replacing the multi-step sampler with a single head evaluation. 

GestureFAR is also practical for lightweight deployment. On a MacBook Air, the one-step student runs at $0.246$ RTF, i.e., about $4.1\times$ faster than real time or approximately $122$ FPS for $30$ FPS motion, showing that continuous token-causal gesture generation can be deployed in real time without high-end GPU hardware.

\begin{table*}[t]
\caption{\textbf{Ablation studies of GestureFAR design choices.}
We isolate representation, generator factorization, distillation objectives, flow-head capacity, distillation scope, and teacher sampling cost.}
\label{tab:ablation-grid}
\vspace{-0.1cm}
\renewcommand{\tabcolsep}{1.8pt}
\small

%----------------------------------------------------
% (a) Continuous VAE vs. discrete tokenization
%----------------------------------------------------
\begin{subtable}[!t]{0.3\linewidth}
\centering
\scalebox{0.84}{
\begin{tabular}{lcc}
\toprule
{\it Representation} & Recon.$\downarrow$ & rFGD$\downarrow$  \\
\midrule
Discrete VQ tokens & 0.121 & 0.213 \\
Discrete RQ tokens & 0.089 & 0.108 \\
Discrete part VQ & 0.075 & 0.053 \\
Continuous VAE & 0.013 & \textbf{0.012} \\
\rowcolor{oursgreen}
\textbf{Causal continuous VAE} & \textbf{0.012} & 0.013 \\
\bottomrule
\end{tabular}}
\caption{\small{Representation}}
\label{tab:ab-representation}
\end{subtable}
\hspace{\fill}
%----------------------------------------------------
% (b) Diffusion forcing vs. discrete autoregression vs. GestureFAR
%----------------------------------------------------
\begin{subtable}[!t]{0.34\linewidth}
\centering
\scalebox{0.86}{
\begin{tabular}{lccc}
\toprule
{\it Generator} & FGD$\downarrow$ & BC$\rightarrow$ & Div.$\rightarrow$ \\
\midrule
Diffusion forcing & 0.612 & 0.792 & 14.26 \\
Discrete autoregression & 0.557 & 0.635 & 13.84 \\
~~~~with RQ generation & 0.483 & 0.764 & \textbf{13.02} \\
~~~~with part VQ & 0.457 & 0.767 & 16.77 \\
\rowcolor{oursgreen}
\textbf{GestureFAR} & \textbf{0.317} & \textbf{0.734} & 13.31 \\
\bottomrule
\end{tabular}}
\caption{\small{Generator factorization}}
\label{tab:ab-generator}
\end{subtable}
\hspace{\fill}
%----------------------------------------------------
% (c) Distillation components
%----------------------------------------------------
\begin{subtable}[!t]{0.32\linewidth}
\centering
\scalebox{0.86}{
\begin{tabular}{lccc}
\toprule
{\it Recipe} & FGD$\downarrow$ & BC$\rightarrow$ & Div.$\rightarrow$ \\
\midrule
DMD & 0.395 & 0.769 & 13.36 \\
ReFlow & 0.424 & 0.677 & 14.23 \\
dCM & 0.570 & \textbf{0.719} & \textbf{13.08} \\
\rowcolor{oursgreen}
dCM+DMD & \textbf{0.308} & 0.741 & 13.24 \\
Teacher & 0.317 & 0.734 & 13.31 \\
\bottomrule
\end{tabular}}
\caption{\small{One-step distillation components}}
\label{tab:ab-distill-components}
\end{subtable}

\vspace{0.2cm}

%----------------------------------------------------
% (d) Flow matching head size
%----------------------------------------------------
\begin{subtable}[!t]{0.3\linewidth}
\centering
\scalebox{0.86}{
\begin{tabular}{lccc}
\toprule
{\it Flow head} & Params$\downarrow$ & FGD$\downarrow$ & BC$\rightarrow$ \\
\midrule
Tiny & \textbf{1.75M} & 0.432 & 0.751 \\
Small & 3.75M & 0.383 & \textbf{0.712} \\
\rowcolor{oursgreen}
\textbf{Default} & 6.51M & \textbf{0.317} & 0.734 \\
Large & 25.07M & 0.321 & 0.723 \\
\bottomrule
\end{tabular}}
\caption{\small{Head capacity}}
\label{tab:ab-head-size}
\end{subtable}
\hspace{\fill}
%----------------------------------------------------
% (e) Distillation scope
%----------------------------------------------------
\begin{subtable}[!t]{0.34\linewidth}
\centering
\scalebox{0.86}{
\begin{tabular}{lccc}
\toprule
{\it Distillation scope} & Trainable & FGD$\downarrow$ & Cost$\downarrow$ \\
\midrule
Whole model & backbone+head & 0.324 & 4h \\
\rowcolor{oursgreen}
\textbf{Head only} & head & \textbf{0.308} & \textbf{10 min} \\
Teacher & none & 0.317 & 8h \\
\bottomrule
\end{tabular}}
\caption{\small{Distillation scope}}
\label{tab:ab-distill-scope}
\end{subtable}
\hspace{\fill}
%----------------------------------------------------
% (f) Teacher solver cost
%----------------------------------------------------
\begin{subtable}[!t]{0.32\linewidth}
\centering
\scalebox{0.86}{
\begin{tabular}{lccc}
\toprule
{\it Solver} & FGD$\downarrow$ & BC$\rightarrow$ & ms/tok$\downarrow$ \\
\midrule
Teacher, $8$ steps & 0.317 & \textbf{0.734} & 24.4 \\
Student, $4$ steps & \textbf{0.304} & 0.768 & 16.2 \\
Student, $2$ steps & 0.308 & 0.744 & 11.9 \\
\rowcolor{oursgreen}
Student, $1$ step & 0.308 & 0.741 & \textbf{9.3} \\
\bottomrule
\end{tabular}}
\caption{\small{Solver cost}}
\label{tab:ab-teacher-steps}
\end{subtable}

\vspace{-5mm}
\end{table*}

\subsection{Ablation Studies}
\label{sec:exp:ablations}

We conduct ablation studies to verify the key designs of GestureFAR, including motion representation, generator factorization, distillation objective, flow-head capacity, distillation scope, and sampling cost with results summarized in Table~\ref{tab:ablation-grid}.

\paragraph{Effect of motion representation.}
We first compare different motion tokenizers. Discrete VQ tokens show a clear reconstruction gap, and stronger variants such as residual quantization and part-based VQ reduce but do not remove this gap. This suggests that discrete tokenization introduces a representation ceiling for high-dimensional full-body motion. The continuous VAE substantially improves reconstruction quality, while the causal continuous VAE keeps comparable quality and makes the representation streamable. This explains why discrete streaming systems such as MIBURI and LiveGesture can be naturally causal but still struggle to match the quality of continuous methods.

\paragraph{Effect of generator factorization.}
We then compare several causal generation paradigms. Diffusion Forcing~\cite{chen2024diffusionforcing} provides a continuous autoregressive alternative, but we find it less effective for co-speech gesture generation. A likely reason is that token-level causal diffusion forcing must condition on noisy motion history, making it harder to learn the fine-grained dependency between speech rhythm and body dynamics. Discrete autoregression~\cite{mughal2026miburi,saleem2026livegesture} avoids this issue and benefits from teacher forcing, but remains limited by the discrete representation. GestureFAR combines the advantages of both: it learns causal dependencies from clean motion history, while the flow head predicts a continuous next-latent distribution. This factorization gives the best overall quality.

\paragraph{Effect of distillation objectives.}
We ablate the objectives used for one-step head distillation. Discrete consistency provides a stable initialization but is not sufficient on its own. Continuous or distribution-matching objectives improve the one-step student, and the best performance is obtained when the warm-up is combined with distribution matching. This supports our multi-procedure design: consistency stabilizes the one-step map, while distribution matching better preserves the teacher's conditional sample distribution.

\paragraph{Flow-head Capacity.}
We vary the size of the flow head. A very small head underfits the conditional latent distribution, while increasing the head to the default size improves generation quality. Further enlarging the head brings little additional gain. This indicates that the backbone should carry the sequence-level audio-motion representation, and the flow head only needs moderate capacity for local sampling.

\paragraph{Distillation scope.}
We compare whole-model distillation with our head-only distillation. Distilling the whole model is more expensive and does not improve the final result. In contrast, head-only distillation reuses cached conditioning vectors from the frozen autoregressive backbone, making training much faster while preserving the learned causal representation. This validates the decomposition of GestureFAR into a representation backbone and a local sampling head.

\paragraph{Effect of sampling cost.}
Finally, we study the solver cost. Reducing the number of sampling steps lowers per-token latency but can affect motion quality and alignment. The distilled one-step head achieves the best efficiency-quality trade-off, removing the repeated flow-head evaluations required by the teacher while keeping competitive gesture quality, making it practical for real-time streaming deployment.

\begin{figure}[thb]
\centering
\includegraphics[width=\linewidth]{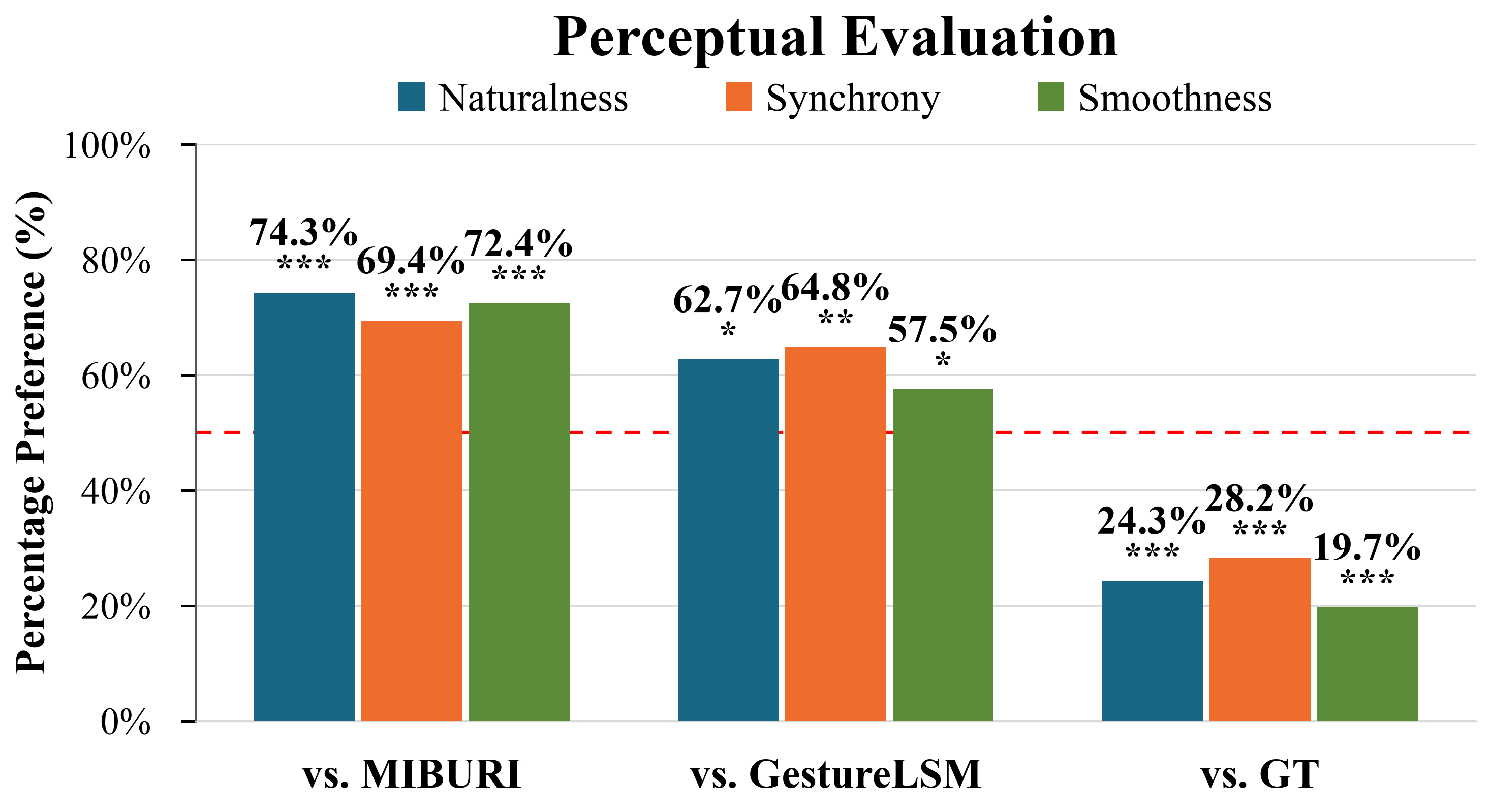}
\vspace{-0.5cm}
\caption{Our GestureFAR have higher user ratings with a clear margin on \textit{Naturalness}, \textit{Synchrony}, and \textit{Smoothness}.}
 \vspace{-0.5cm}
\label{fig:user-study}
\end{figure}

\subsection{User Study}
\label{sec:exp:userstudy}

We conduct a user study to evaluate the perceptual quality of GestureFAR. The study includes $20$ participants and $240$ generated video samples, with $80$ samples from each method: GestureFAR, GestureLSM~\cite{liu2025gesturelsm}, and MIBURI~\cite{mughal2026miburi}. For each participant, videos are presented in a randomized order to reduce ordering bias. Participants rate each video on a $1$--$5$ Mean Opinion Score (MOS) scale, where $1$ denotes the lowest quality and $5$ denotes the highest quality. We evaluate three perceptual criteria: \textit{realness}, speech-gesture \textit{synchrony}, and motion \textit{smoothness}. \textbf{Realness} measures whether the generated gestures resemble natural human motion in pose, expressiveness, and overall plausibility. \textbf{Synchrony} measures whether gestures are temporally aligned with the speech rhythm and audio content. \textbf{Smoothness} measures motion continuity, penalizing abrupt stops, unnatural jerks, and poor whole-body coordination. As shown in Fig.~\ref{fig:user-study}, GestureFAR achieves the highest MOS across all three criteria. The results indicate that GestureFAR produces gestures that are perceived as more realistic, better synchronized with speech, and smoother than those generated by the competing methods.

\section{Conclusion}
\label{sec:conclusion}

In this paper, we presented \textbf{GestureFAR}, a streaming token-causal framework for co-speech gesture generation. GestureFAR moves beyond discrete autoregressive gesture tokens by autoregressing over continuous motion latents with a flow-autoregressive design, preserving both streaming causality and continuous motion expressiveness. To make the framework practical for live deployment, we further proposed Multi-Procedure Distribution Matching Distillation, which freezes the causal backbone and distills only the multi-step flow head into a one-step sampler. We hope GestureFAR can serve as a step toward realistic, low-latency, and deployable gesture generation for real-time conversational avatars.

\appendix

\appendix
\newpage

\begin{center} 
    \centering
    \textbf{\large GestureFAR: Streaming Co-Speech Gesture Generation with Flow Autoregression}
\end{center}
\begin{center} 
    \centering
    \large Supplementary Material
\end{center}

\section{One-Step Distillation: Formal Details}\label{app:distill}
This appendix gives the parametrization and objectives summarized in the main paper. Throughout, $\mathbf{c}$ is a frozen conditioning vector harvested from the teacher backbone, and all quantities live in the $\sigma_\text{data}\!=\!1$ normalized latent space of the flow head.

\paragraph{TrigFlow parametrization.}
We map the rectified-flow time $\tau$ to the TrigFlow angle $s=\arctan(\sigma)$ with $\sigma=\tau/(1-\tau)$, so the forward process is $\mathbf{x}_s=\cos s\,\mathbf{x}_0+\sin s\,\boldsymbol{\epsilon}$ ($s\!=\!0$ data, $s\!=\!\tfrac{\pi}{2}$ noise). The head outputs a TrigFlow velocity $F_\theta(\mathbf{x}_s,s,\mathbf{c})$, from which EDM-style preconditioning recovers the clean prediction $\hat{\mathbf{x}}_0=\cos s\,\mathbf{x}_s-\sin s\,F_\theta$. Because $\sigma_\text{data}\!=\!1$ the preconditioning coefficients reduce to these trigonometric forms; we verify $\mathrm{std}(\mathbf{x}_0)\!\approx\!1$ on the harvested cache and rescale otherwise.

\paragraph{Discrete warm-up (dCM).}
Given $\mathbf{x}_s$ at a sampled $s$, we integrate the \emph{teacher}'s velocity with a few Euler steps to a cleaner angle $s'\!<\!s$ and read its clean prediction $\hat{\mathbf{x}}_0^\text{tea}(s')$. The student is trained to match this stop-gradient target in one shot, $\mathcal{L}_\text{dCM}=\|\hat{\mathbf{x}}_0^\theta(\mathbf{x}_s,s)-\mathrm{sg}[\hat{\mathbf{x}}_0^\text{tea}(s')]\|^2$. This needs no network derivatives and is numerically stable, but the target inherits the teacher solver's discretization error, so we use dCM only to initialize.

\paragraph{Distribution matching (DMD).}
DMD adds a reverse-KL term through a third ``fake-score'' network $s_\psi$, trained to denoise the \emph{student}'s own one-step samples (a flow-matching loss on $\hat{\mathbf{x}}_0^\theta$). The generator gradient pushes the student toward the teacher's score,
\begin{equation}
\nabla_\theta\mathcal{L}_\text{DMD}=\mathbb{E}_{s}\big[(\hat{\mathbf{x}}_0^{\,\text{fake}}-\hat{\mathbf{x}}_0^{\,\text{tea}})\,\partial_\theta \hat{\mathbf{x}}_0^\theta\big],
\label{eq:app-dmd}
\end{equation}
evaluated on the student's one-step output re-noised at $s$. Following~\cite{zheng2026rcm} we update critic and generator in alternation and sum the DMD term with the dCM loss; all DMD runs use a fixed weight. Classifier-free guidance enters through the teacher score $\hat{\mathbf{x}}_0^\text{tea}$, combining its conditional and unconditional predictions at a fixed scale, which bakes guidance into the student.

\section{Additional Experiments}
\label{sec:sup-exp}

\paragraph{Multi-speaker Setting.}
In the main paper, we show the performance of single speaker, here we show the multi-speaker setting. Table~\ref{tab:sup-experiment} presents the quantitative results on the BEAT-2 benchmark. Our model, \textbf{GestureFAR}, achieves state-of-the-art performance across all key metrics. Notably, our method obtains the lowest FGD (\textbf{0.289}), indicating the highest overall realism, while maintaining strong beat consistency (0.484) and natural motion diversity (7.63). These results demonstrate the benefit of our intentional alignment and conditioning mechanisms in generating gestures that are both semantically expressive and rhythmically precise.

\begin{table}
    \caption{The quantitative results on BEAT-2 all speaker setting. We bold the best results.}
    \label{tab:sup-experiment}
    \centering
    \resizebox{\linewidth}{!}{
    \begin{tabular}{l|cccccc} 
    \toprule
    Methods & FGD ($\downarrow$) & BC ($\rightarrow$) & Diversity ($\rightarrow$)\\
    \midrule
    Ground-Truth & -- & 0.477 & 7.29\\
    \midrule
    
    $\text{CaMN}$~\cite{liu2022beat} & 0.512    & 0.200  & 5.58\\
    $\text{EMAGE}$~\cite{liu2023emage} & 0.692 & 0.284 & 6.06 \\    
    GestureLSM~\cite{liu2025gesturelsm}  & 0.466 & 0.525 & 9.23 \\
    \midrule 
    $\text{MIBURI}$~\cite{mughal2026miburi} & 0.480 & 0.461 & 10.44 \\
    \rowcolor{mygray} GestureFAR  & \textbf{0.289}  & \textbf{0.484} & 7.63 \\
    \bottomrule
    \end{tabular}
    }
\end{table}

\paragraph{Results on Audio2PhotoReal.}
In addition to the experiments on BEAT2 as shown in the main paper, we present the quantitative results on the Audio2PhotoReal~\cite{ng2024audio2photoreal} benchmark in
Table~\ref{tab:sup-experiment2}. Our model, \textbf{GestureFAR}, achieves state-of-the-art performance across all key metrics. These results demonstrate the strong capability of model to get generalized to dyadic conversational speaking and listening settings.

\begin{table}
    \caption{The quantitative results on Audio2PhotoReal. We bold the best results.}
    \label{tab:sup-experiment2}
    \centering
    \resizebox{\linewidth}{!}{
    \begin{tabular}{l|cccccc} 
    \toprule
    Methods & FGD ($\downarrow$) &  Diversity ($\rightarrow$)\\
    \midrule
    Ground-Truth & -- & 2.50\\
    \midrule
    $\text{EMAGE}$~\cite{liu2023emage} & 4.43 & 2.13 \\
    $\text{Audio2PhotoReal}$~\cite{ng2024audio2photoreal} & 2.94 & 2.36 \\
    GestureLSM~\cite{liu2025gesturelsm}  & 2.64 & 2.34 \\
    MIBURI~\cite{mughal2026miburi}  & 5.12 & 2.01 \\
    \midrule 
    \rowcolor{mygray} GestureFAR  & \textbf{2.08} & \textbf{2.51} \\
    \bottomrule
    \end{tabular}
    }
\vspace{-0.5cm}
\end{table}

\section{Metric Details}

\paragraph{Fr\'{e}chet Gesture Distance (FGD)}
Fr\'{e}chet Gesture Distance (FGD), introduced in \cite{yoon2020speech}, quantifies the similarity between the distributions of real and generated gestures, where a lower FGD signifies a closer match. Inspired by perceptual loss in image generation, FGD is computed using latent features extracted from a pretrained network:
\begin{equation}
\label{eqfid}
\resizebox{.85\hsize}{!}{$
\operatorname{FGD}(\mathbf{g}, \hat{\mathbf{g}})=\left\|\mu_{r}-\mu_{g}\right\|^{2}+\operatorname{Tr}\left(\Sigma_{r}+\Sigma_{g}-2\left(\Sigma_{r} \Sigma_{g}\right)^{1 / 2}\right),$}
\end{equation}
where $\mu_{r}$ and $\Sigma_{r}$ denote the mean and covariance of the latent feature distribution $z_{r}$ derived from real gestures $\mathbf{g}$, while $\mu_{g}$ and $\Sigma_{g}$ correspond to the statistics of the generated gestures $\hat{\mathbf{g}}$.

\paragraph{L1 Diversity}
L1 Diversity, proposed in \cite{li2021audio2gestures}, measures the variation across multiple gesture sequences, with higher values indicating greater diversity. The average L1 distance across $N$ motion sequences is computed as:
\begin{equation}
\resizebox{.75\hsize}{!}{$
    \text{L1 div.} =  \frac{1}{2 N (N-1)} \sum_{t=1}^{N} \sum_{j=1}^{N} \left\|p_{t}^{i}-\hat{p}_{t}^{j}\right\|_{1},$}
\end{equation}
where $p_{t}$ represents the joint positions at frame $t$. Diversity is evaluated on the complete test set. To ensure a focus on local motion, global translation is neutralized when computing joint positions.

\paragraph{Beat Constancy (BC)}
Beat Constancy (BC), as defined in \cite{li2021ai}, assesses the temporal alignment between gestures and audio rhythm. Higher BC values indicate stronger synchronization. Speech onsets are treated as audio beats, while motion beats correspond to local minima in the upper body joint velocity (excluding fingers). The alignment is determined using:
\begin{equation}
\label{align}
\resizebox{.80\hsize}{!}{$
\text{BC}= \frac{1}{g} \sum_{b_{g}\in g} \exp \left(-\frac{\min _{b_{a}\in a}\left\|b_{g}-b_{a}\right\|^{2}}{2 \sigma^{2}}\right),$}
\end{equation}
where $g$ and $a$ denote the sets of detected gesture beats and audio beats, respectively.

% References and End of Paper
\bibliography{main}

\end{document}